\documentclass[default,iicol,sn-vancouver,Numbered]{sn-jnl}

\usepackage{graphicx}%
\usepackage{multirow}%
\usepackage{amsmath,amssymb,amsfonts}%
\usepackage{amsthm}%
\usepackage{mathrsfs}%
\usepackage[title]{appendix}%
\usepackage{xcolor}%
\usepackage{textcomp}%
\usepackage{manyfoot}%
\usepackage{booktabs}%
\usepackage{listings}%

\usepackage[linesnumbered]{algorithm2e}
\RestyleAlgo{ruled}
\SetKwComment{Comment}{/* }{ */}
\SetKwInput{KwRequire}{Require}

\usepackage{listings}
\usepackage{xcolor}
\usepackage{changepage}
\usepackage{enumitem}

\lstdefinelanguage{PDDL}{
    morekeywords={define,domain,requirements,strips,typing,types,predicates,action,parameters,precondition,effect,and,not,or,forall,when,:either,:domain,:requirements,:types,:predicates,:action,:parameters,:precondition,:effect},
    morecomment=[l]{;},
    morecomment=[s]{/*}{*/},
    morestring=[b]",
}

\newcommand{\huang}{LMZSP}
\newcommand{\prog}{ProgPrompt}
\newcommand{\inner}{Inner Monologue}
\newcommand{\num}{1,085}

\begin{document}

\title[Article Title]{Integrating Action Knowledge and LLMs for Task Planning and Situation Handling in Open Worlds}


\author[1]{\fnm{Yan} \sur{Ding}}
\author[1]{\fnm{Xiaohan} \sur{Zhang}}
\author[1]{\fnm{Saeid} \sur{Amiri}}
\author[1]{\fnm{Nieqing} \sur{Cao}}
\author[2]{\fnm{Hao} \sur{Yang}}
\author[2]{\fnm{Andy} \sur{Kaminski}}
\author[2]{\fnm{Chad} \sur{Esselink}}
\author[1]{\fnm{Shiqi} \sur{Zhang}}
\affil[1]{\orgdiv{The State University of New York at Binghamton}, \orgaddress{\state{NY} \postcode{13902}, \country{USA}}}
\affil[2]{\orgdiv{Ford Motor Company}, \orgaddress{\state{MI} \postcode{18900}, \country{USA}}}


\abstract{    
Task planning systems have been developed to help robots use human knowledge (about actions) to complete long-horizon tasks.
Most of them have been developed for ``closed worlds'' while assuming the robot is provided with complete world knowledge.
However, the real world is generally open, and the robots frequently encounter unforeseen situations that can potentially break the planner's completeness. 
Could we leverage the recent advances on pre-trained Large Language Models (LLMs) to enable classical planning systems to deal with novel situations? \\

This paper introduces a novel framework, called COWP, for open-world task planning and situation handling. 
COWP dynamically augments the robot's action knowledge, including the preconditions and effects of actions, with task-oriented commonsense knowledge. 
COWP embraces the openness from LLMs, and is grounded to specific domains via action knowledge. 
For systematic evaluations, we collected a dataset that includes \num~execution-time situations. 
Each situation corresponds to a state instance wherein a robot is potentially unable to complete a task using a solution that normally works.
Experimental results show that our approach outperforms competitive baselines from the literature in the success rate of service tasks.
Additionally, we have demonstrated COWP using a mobile manipulator.
Supplementary materials are available at: 
\url{https://cowplanning.github.io/}
}




\maketitle

\begin{figure*}[htp]
\vspace{1em}
\centering
\includegraphics[width=2.0\columnwidth]{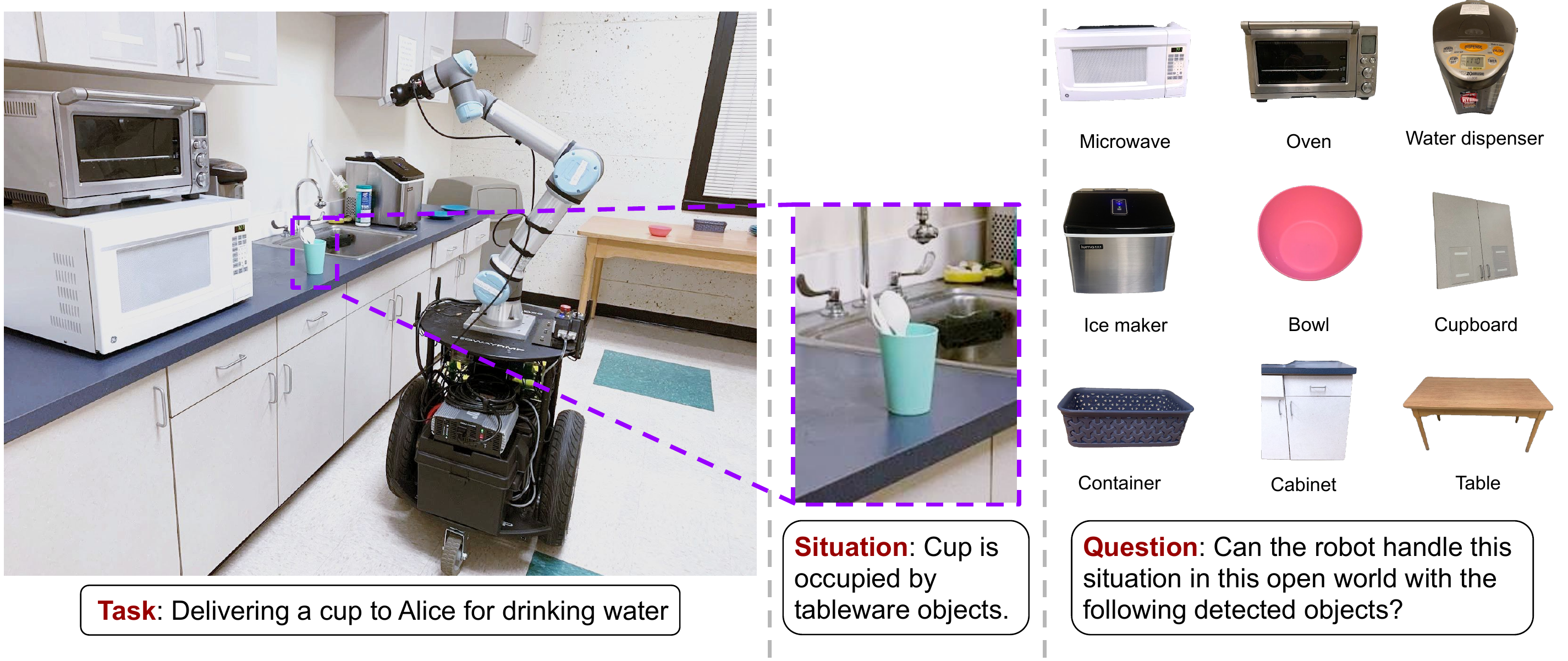}
\caption{An illustrative example of a \textit{situation} in the real world, encountered during the execution of the plan ``delivering a cup to a human for drinking water.'' 
The robot approached a cabinet in a kitchen room on which a cup was located.
The robot then found the cup to be delivered. 
Before grasping it, however, the robot detected a situation that \textit{the cup was occupied with a fork, a knife, and a spoon}. 
This situation prevented the robot from performing the current action (i.e., grasping) and rendered normal solutions for drinking water invalid. 
On the left is a kitchen environment and our mobile manipulator detecting a cup. 
On the right, we show a few visually perceived objects that were potentially useful for situation handling. 
}\label{fig:mobile_manipulator}
\end{figure*}

\section{Introduction}
Robots that operate in the real world frequently encounter long-horizon tasks that require multiple actions.
Automated task planning algorithms have been developed to help robots sequence actions to accomplish those tasks~\cite{ghallab2016automated}.
Closed world assumption (CWA) is a presumption that was developed by the knowledge representation community and states that ``statements that are true are also known to be true''~\cite{reiter1981closed}. 
Most current task planners have been developed for closed worlds, assuming complete domain knowledge is provided and one can enumerate all possible world states~\cite{knoblock1991characterizing, hoffmann2001ff, nau2003shop2, helmert2006fast}.
However, the real world is ``open'' by nature, and unforeseen situations are common in practice~\cite{hanheide2017robot}. 
As a consequence, current automated task planners that are largely knowledge-based tend to be fragile in open worlds rife with unforeseen situations.
Fig.~\ref{fig:mobile_manipulator} shows an example situation:  \emph{Aiming to grasp a cup for drinking water, a robot found that the cup was not empty}. 
Although one can name many such situations, it is impossible to provide a complete list of them. 
To this end, researchers have developed open-world planning methods for robust task completions in real-world scenarios~\cite{jiang2019open,chernova2020situated,hanheide2017robot,kant2022housekeep,huang2022language,brohan2023can}.

In the current literature, there are at least three ways of addressing the open-world planning problem. 
The first involves acquiring knowledge via human-robot interaction, e.g., dialog-based, to handle situations in an open-world context~\cite{perera2015learning, amiri2019augmenting,tucker2020learning}.
Those methods require human involvement, which might hinder their autonomy capability and limit their applicability in the real world. 
A second idea for open-world planning relies on dynamically building a knowledge base to assist a pre-defined task planner, where this knowledge base is usually constructed in an automatic way using external information, e.g., using open-source knowledge graphs~\cite{jiang2019open,chernova2020situated,hanheide2017robot}. 
Such external knowledge bases are considered ``bounded'' in this paper due to their representation and knowledge source, which limits the ``openness'' of their task planners. 
Large Language Models (LLMs) have been developed in recent years~\cite{brown2020language,zhang2022opt,openai,google-bard-faq}, and those LLMs have demonstrated major improvements in a variety of downstream tasks. 
Researchers have investigated the idea of extracting common sense from LLMs to guide robot task planning~\cite{brohan2023can, huang2022language, elsweiler2022food, kant2022housekeep}. 
One challenge in this process is that the knowledge from LLMs is \emph{domain-independent}~\cite{davis2015commonsense, huang2023grounded}, whereas the robot faces specific domains that are featured with many \emph{domain-dependent} constraints. 
For example, an LLM may provide a robot with the knowledge of how to serve water to people, but it falls short in determining the available types of containers (such as cups or glasses) the robot has access to, as well as the water sources in stock (like tap water or bottled water). 
In line with the third idea, we use LLMs for open-world planning in this paper. 
To address the challenge of grounding common sense to specific domains, we propose to enable the marriage of classical AI planning methods, and LLM-based knowledge extraction.

In this paper, we develop a robot task planning and situation handling framework, called \textit{Common sense-based Open-World Planning}~(\textbf{COWP}), that uses an LLM for dynamically augmenting automated task planners with external task-oriented common sense.
COWP is based on classical planning and leverages LLMs to augment action knowledge (action preconditions and effects) for task planning and situation handling. 
The \textbf{main contribution} of this work is a novel integration of a pre-trained LLM with a knowledge-based task planner. 
Inheriting the desirable features from both sides, COWP is well grounded in specific domains while embracing commonsense solutions at large. 

To conduct systematic evaluations, we selected 12 dining-focused tasks from the \emph{ActivityPrograms} dataset~\cite{puig2018virtualhome}. 
Each task consisted of a high-level name, such as ``Serve water'', and a natural language description of the action sequence required to complete it. 
Using a crowdsourcing platform, we recruited 112 participants to propose potential situations that might hinder successful task execution. 
We gathered a situation dataset that includes more than one thousand situations for evaluation purposes. 
According to experimental results, we see COWP performed better than literature-selected baselines~\cite{jiang2019open,huang2022language,singh2023progprompt} in terms of the respective success rates in task completion and situation handling.
We implemented and demonstrated COWP using a mobile manipulator. 

\section{Background and Related Work}
In this section, we first briefly discuss classical task planning methods that are mostly developed under the closed world assumption. 
We then summarize three families of open-world task planning methods for robots, which are grouped based on how unforeseen situations are addressed. 

\vspace{.5em}
\noindent
\textbf{Classical Task Planning for Closed Worlds:}
A classical task planning problem consists of two main components: a domain description and a problem description~\cite{haslum2019introduction}. 
The domain description includes a collection of actions, each of which is defined by its preconditions and subsequent effects. 
The problem description, on the other hand, specifies the initial state and the desired goal conditions. 
A sequence of actions can be generated, enabling an effective transition from the initial state to the designated goal state.

Closed world assumption (CWA) indicates that an agent is provided with complete domain knowledge, and that all statements that are true are known to be true by the agent~\cite{reiter1981closed}. 
In this paper, such a classical planning system is referred to as a closed-world task planner. 
Although robots face the real world that is open by nature, their planning systems are frequently constructed under the CWA~\cite{hanheide2017robot,jiang2019task,galindo2008robot,ghallab2016automated,haslum2019introduction,nau2003shop2}. 
The consequence is that those robot planning systems are not robust to unforeseen situations at execution time. 
In this paper, we aim to develop a task planner that is aware of and able to handle unforeseen situations in open-world scenarios.

\vspace{.5em}
\noindent
\textbf{Open-World Task Planning with Human in the Loop:} 
Task planning systems have been developed to acquire knowledge via human-robot interaction to handle open-world situations~\cite{perera2015learning, amiri2019augmenting, tucker2020learning}. 
For instance, researchers created a planning system that uses dialog systems to augment their knowledge bases~\cite{perera2015learning}, whereas Amiri et al. (2019) further modeled the noise in language understanding~\cite{amiri2019augmenting}.
Tucker et al. (2020) enabled a mobile robot to ground new concepts using visual-linguistic observations, e.g., to ground the new word ``box'' given command of ``move to the box'' by exploring the environment and hypothesizing potential new objects from natural language~\cite{tucker2020learning}. 
The major difference from those open-world planning methods is that COWP does not require human involvement. 

\vspace{.5em}
\noindent
\textbf{Open-World Task Planning with External Knowledge:}
Some existing planning systems address unforeseen situations by dynamically constructing an external knowledge base for open-world reasoning.
For instance, researchers have developed object-centric planning algorithms that maintain a database about objects and introduce new object concepts and their properties (e.g., location) into their task planners~\cite{jiang2019open, chernova2020situated}.
For example, Jiang~et~al.~(2019) developed an object-centric, open-world planning system that dynamically introduces new object concepts through augmenting a local knowledge base with external information~\cite{jiang2019open}.
In the work of Hanheide et al. (2017), additional action effects and assumptive actions were modeled as an external knowledge to explain the failure of task completion and compute plans in open worlds~\cite{hanheide2017robot}.
A major difference from their methods is that COWP employs an LLM as a generative approach that is capable of responding to any situation, whereas the external knowledge sources of those methods limits the openness of their systems. 

\vspace{.5em}
\noindent
\textbf{Closed-World Task Planning with LLMs:}
A straightforward idea for LLM-based planning is to directly enter planning domain information into LLMs, and request plan generations. 
Recent research has shown that a naive implementation of such ideas produces very poor performance even if the planning domain is as simple as Blocksworld~\cite{valmeekam2022large,valmeekam2023planning}. 
The ChatGPT report (in its conclusion section) also identified ``long-horizon planning'' as a challenging task, and encouraged more research on LLM-based planning~\cite{openai2023gpt4}. 
Very recently, researchers have investigated translating descriptions of planning tasks in natural language into PDDL~\cite{haslum2019introduction}, and then computing plans using PDDL systems~\cite{liu2023llm}. 
The produced system called LLM+P takes natural language descriptions as input, and computes plans with optimality guarantee. 
The action knowledge of LLM+P formulated in PDDL was provided by domain experts, and could not adapt to novel situations at execution time. 
By comparison, COWP (ours) dynamically extract common sense from LLMs for augmenting its action knowledge for planning and situation handling. 

\vspace{.5em}
\noindent
\textbf{Open-World Task Planning with LLMs:}
In recent years, many LLMs have emerged, including BERT~\cite{devlin2018bert}, GPT-3~\cite{brown2020language}, ChatGPT~\cite{openai}, CodeX~\cite{chen2021evaluating}, and OPT~\cite{zhang2022opt}. 
These LLMs encode a large amount of commonsense knowledge~\cite{liu2023pre,wang2021can,li2022pre,west2021symbolic} and have been employed in robot task planning~\cite{kant2022housekeep,huang2022language,brohan2023can,huang2022inner,singh2023progprompt,ding2023task}. 
For instance, the work of Huang et. al.~\cite{huang2022language} showed that LLMs can effectively facilitate task planning in household environments by iteratively augmenting prompts~\cite{huang2022language}. 
The SayCan system enabled robot planning with affordance functions to account for action feasibility, where the service requests are specified in natural language (e.g., ``deliver a Coke'')~\cite{brohan2023can}. 
Additionally, various teams also have successfully applied LLMs to generate plans for high-level tasks expressed in natural language by sequencing actions~\cite{huang2022inner,kant2022housekeep,xie2023translating,song2022llm,lin2023text2motion,ding2023task}. 

It is generally difficult to ground LLM-based planning systems in the real world as the LLMs were not trained for specific domains. 
For instance, LLM-based planners frequently generate plans that require objects or tools that are not present in the scene~\cite{huang2022inner}. 
COWP (ours) alleviates this issue through reasoning about robot skills using rule-based action knowledge. 

Work closest to COWP is a recent LLM-based planning system, called~\prog~\cite{singh2023progprompt}, which generates task plans and handles situations using programmatic LLM prompts. 
\prog~handles situations by asserting preconditions of the plan (e.g., being close to the fridge before attempting to open it) and responding to failed assertions with appropriate recovery actions. 
Compared with~\prog, which relies on example solutions in prompting to guide the LLMs, COWP (ours) uses action knowledge to enable zero-shot prompting for planning and situation handling.
For example, in the case of~\prog, three examples are typically provided in the prompt. 
These examples guide~\prog to generate task plans, and these plans take into account feedback from the environment by including preconditions for actions. 
By comparison, COWP makes use of action knowledge, which can be in the form of rules, facts, and principles related to the task. 
This can include a detailed understanding of how actions impact the state of the environment, which is not required in the prompts used by~\prog. 
This enables COWP to create task plans without the need for example solutions, i.e., zero-shot prompting.


COWP extracts commonsense knowledge from LLMs and incorporates rule-based action knowledge from human experts. 
Reasoning with action knowledge ensures the soundness of task plans generated by COWP, while querying LLMs guarantees the openness of COWP to unforeseen situations. 
As a result, COWP can be better grounded to specific domains, and is able to incorporate common sense to augment robot capabilities supported by predefined skills.


\begin{figure*}[h]
\centering
\includegraphics[width=2.1\columnwidth]{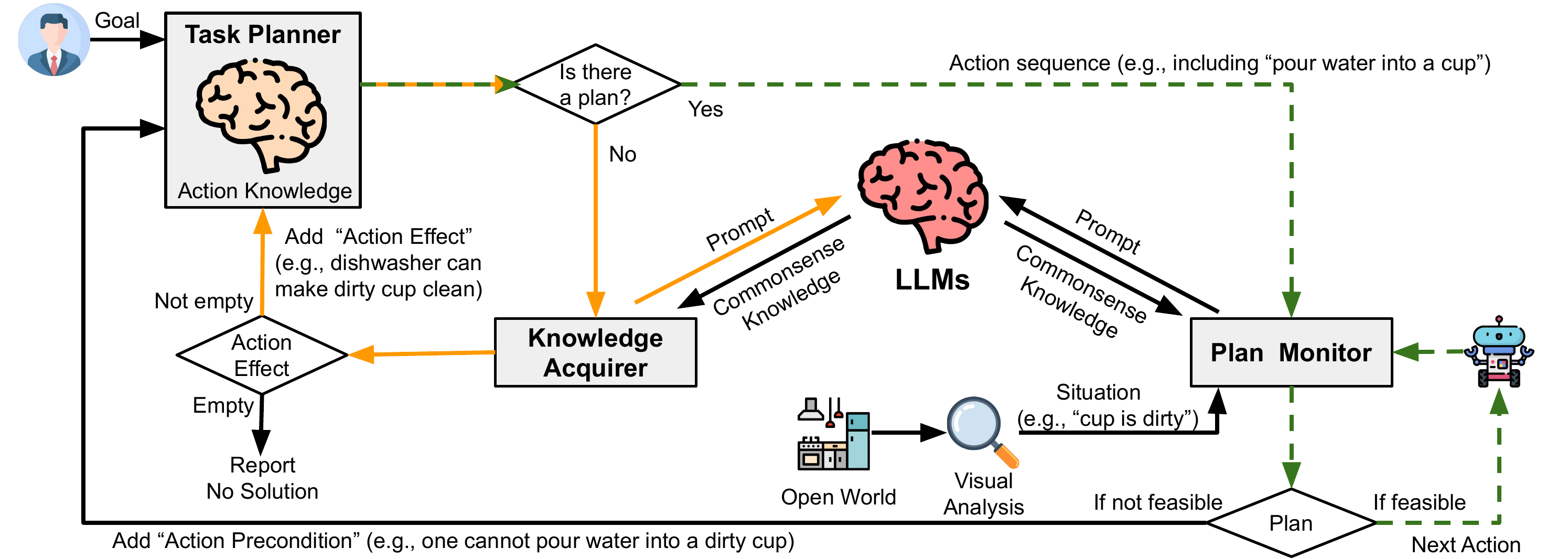}
\vspace{0.5em}
\caption{An overview of \textbf{COWP} that includes the three key components of Task Planner (provided as prior knowledge under closed-world assumption), Knowledge Acquirer, and Plan Monitor. 
The \textbf{\color{green!50!black} green} (dashed) loop represents a plan execution process where the robot does not encounter any situation, or these situations have no impact on the robot's plan execution. 
The \textbf{\color{orange} orange} loop is activated when the robot's current (closed-world) task planner is unable to develop a plan, which activates Knowledge Acquirer to augment the task planner with additional action effects utilizing common sense. 
%
}
\label{fig:overview}
\end{figure*}

\section{Algorithm}
\label{sec:alg}
In this section, we first provide a problem statement and then present our open-world planning approach called Common sense-based Open-World Planning (COWP). 


Our goal is to address open-world planning problems for robots. 
An open-world planning problem assumes that the robot might encounter situations that are not considered in the development of the planning systems. 
More specifically, we use the term of \emph{situation} to refer to an unforeseen world state that potentially prevents an agent from completing a task using a solution that normally works.
In this paper, we assumed the availability of a description of the robot's skills, formulated in action description language PDDL.
PDDL is designed to formalize AI planning problems, allowing for a more direct comparison of planning algorithms and implementations~\cite{aeronautiques1998pddl}.
The \emph{objective} of an open-world planner is to compute plans that can adapt to and handle unexpected situations while pursuing the completion of service tasks or reporting ``no solution'' when appropriate.

\begin{algorithm}
\small
\caption{COWP algorithm}\label{alg:COWP}
\KwRequire{Task Planner, Plan Monitor, Knowledge Acquirer, and an LLM}
\KwIn{A domain description, and a problem description}
\While{\rm task is not completed}
{
    Compute a plan \textit{pln} using Task Planner given the domain description and the problem description\label{l:tp}\;
    \For{\rm each action in \textit{pln}\label{l:for_s}}
    {   
        Evaluate the feasibility of the current plan using Plan Monitor given a situation\label{l:pm_e}\;
        \eIf{\rm a plan is not feasible\label{l:sit}}
        {
                Add an action precondition to Task Planner\label{l:ka_s}\;
                Compute a new task plan $\textit{pln}'$, and \textit{pln} $\leftarrow$ $\textit{pln}'$\label{l:tp'}\;
                \If{\em $\textit{pln}$ is empty\label{l:pln'_empty}}
                {
                    Extract common sense (action effects) using Knowledge Acquirer\label{l:ka}\;
                    \eIf{\em action effect is not empty}
                    {
                        Add an action effect to Task Planner\label{l:ka_effect}\;
                        Compute a new task plan $\textit{pln}''$, and \textit{pln} $\leftarrow$ $\textit{pln}''$\label{l:tp''}\;
                    }
                    {
                        Report ``no solution''\label{l:quit}\;
                    }
                }\label{l:ka_e}
        }
        {
            Execute the next action by the robot\label{l:exe_act}\;
        }
    }\label{l:for_e}
}
\end{algorithm}

\subsection{Algorithm Description}
Fig.~\ref{fig:overview} illustrates the three major components of our COWP framework.
\textbf{Task Planner} is used for computing a plan under the closed-world assumption and is provided as prior knowledge in this work. 
\textbf{Plan Monitor} evaluates the overall feasibility of the current plan using common sense. 
\textbf{Knowledge Acquirer} is for acquiring common sense to augment the robot's action effects when the task planner generates no plan.

Algorithm~\ref{alg:COWP} describes how the components of COWP interact with each other. 
Initially, Task Planner generates a satisfying plan based on the goal provided by a human user in \textbf{Line~\ref{l:tp}}. 
After that, the actions in the plan are performed sequentially by a robot in the for-loop of \textbf{Lines~\ref{l:for_s}-\ref{l:for_e}}.
If the current plan remains feasible, as evaluated by Plan Monitor, the next action will be directly passed to the robot for execution in \textbf{Line~\ref{l:exe_act}}; 
otherwise, an action precondition will be added to Task Planner in \textbf{Line~\ref{l:ka_s}}. 
For example, when Task Planner does not know anything about ``dirty cups,'' it might wrongly believe that one can use a dirty cup as a container for drinking water. 
In this situation, \textbf{Line~\ref{l:ka_s}} will add a new statement ``\emph{The precondition of filling a cup with water is that the cup is not dirty}''  into the task planner. 
After the domain description of Task Planner is updated (adding action preconditions), the planner tries to  generate a new plan $\textit{pln}'$ in \textbf{Line~\ref{l:tp'}}. 
If no plan is generated by Task Planner (\textbf{Line~\ref{l:pln'_empty}}), Knowledge Acquirer will be activated for knowledge augmentation with external task-oriented common sense in \textbf{Line~\ref{l:ka}}. 
If the extracted common sense includes action effects, such information will be added into the task planner in \textbf{Line~\ref{l:ka_effect}}. 
For instance, the robot might learn that a chopping board can be used for holding steak, as an action effect that was unknown before.
This process continues until no additional action effects can be generated (in this case, COWP reports ``no solution'' in \textbf{Line~\ref{l:quit}} or a plan is generated.


COWP leverages common sense to augment a knowledge-based task planner to address situations towards robust task completion in open worlds. 
The implementations of \textbf{Lines~\ref{l:pm_e}} and \textbf{\ref{l:ka}} using common sense are non-trivial in practice. 
Next, we describe how commonsense knowledge is extracted from an LLM for those purposes.

\subsection{Plan Monitor and Knowledge Acquirer}\label{sec:pm_ka}
In this section, two important components of COWP are discussed, i.e., Plan Monitor and Knowledge Acquirer.

\textbf{Plan Monitor} is designed to evaluate if there are any situations that could prevent a robot from completing its current task successfully. 
To achieve this, the plan monitor takes in action sequences generated from a classical planner and a set of situations collected from the real world. 
The Plan Monitor uses a prompt template to query an LLM for each action in the sequence. 
The prompt template is based on the following structure:


\vspace{.5em}
\begin{flushleft}
    Prompt 1: \emph{Is it suitable for a robot to [Perform-Action], if [Situation]?} 
\end{flushleft}

\vspace{.5em}

The placeholder [] represents the specific information to be filled in within the template. 
In this case, \emph{[Perform-Action]} represents a natural language description of an action generated from our PDDL-based task planner, with the action in the form of ``Action Object-1 Object-2 ...''.
\emph{[Situation]} represents a natural language description of a situation.
For example, if the current action is ``Fill Cup Water'' and the given situation is ``Cup is broken'', the corresponding prompt would be ``\emph{Is it suitable for a robot to fill a cup with water, if the cup is broken?}''.
Translating a symbolic action into a natural language description can be achieved either by following handcrafted rules or with the assistance of LLM's grammar completion.
In this case, we adopt the first approach.
An LLM generates a natural language response to each prompt. If the LLM determines that the action is infeasible given the situation described, the whole plan is considered infeasible.

\textbf{Knowledge Acquirer} aims to acquire common sense for augmenting the task planner's action knowledge for situation handling. 
In particular, the input for the knowledge acquirer consists of a collection of available objects in the environment that the robot can access. 
Another template-based prompt is developed for querying an LLM for acquiring common sense about action effects.


\vspace{.5em}
\begin{flushleft}
    Prompt 2: \emph{Is it suitable for a robot to [Perform-Action-with-Object]?}\footnote{Intuitively, the solution from COWP goes beyond finding alternative objects in addressing unforeseen situations. COWP enables situation handling by manipulating the attributes of individual instances. For example, a ``dirty cup'' situation can be handled by running a dishwasher, where no second object is involved.}
\end{flushleft}  
\vspace{.5em}


In the prompt, the placeholder \emph{[Perform-Action-with-Object]} represents a natural language description of a robot performing an action with another object.
For example, if the action ``Fill Cup Water'' is considered infeasible under the situation ``Cup is broken'' by Plan Monitor, one instance of using Prompt 2 with another object ``bowl'' is ``\emph{Is it suitable for a robot to fill a bowl with water?}''. 
If the response from an LLM to the example prompt is ``\emph{Yes}'', an additional action effect, ``The bowl can also be a container to fill water'', will be added to the task planner.

Continuing the ``Serve water'' example, additional action effects introduced through Prompt 2 might enable the task planner to generate many feasible plans, e.g., using a glass, measuring cup, or bowl to fill water.
It can be difficult for the task planner to evaluate which plan makes the best sense. 
To this end, we develop Prompt 3 for selecting a task plan of the highest quality among those satisficing plans:




\vspace{.5em}
\begin{flushleft}
    Prompt 3: \emph{There are some objects, such as  [Object-1], [Object-2], ..., and [Object-N]. Which is the most suitable for [Current-Task], if [Situation]?}
\end{flushleft}
\vspace{.5em}

The placeholder \emph{[Current-Task]} is a high-level task description in natural language.
In the example of ``Serve water'', we expect the LLM to respond by suggesting ``glass'' being the most suitable for holding water among those mentioned items. 



\subsection{Implementation}\label{sec:pm_ka_example}

Here, we explain how to implement COWP using the ``Serve water'' task as an example. 
To do so, we use a PDDL-based closed-world task planner, which requires both a domain file and a problem file.
The domain file defines a set of predicates (e.g., \texttt{is\_grasped}) and actions (e.g., \emph{fill}), with each action specified by its preconditions and effects. 
Consider the following action definition for \emph{fill} shown in Fig.~\ref{fig:fill1}, which includes preconditions like \texttt{(is\_grasped ?c) $\wedge$ (is\_empty ?c)} and effects such as \texttt{(is\_filled ?c)}:

\begin{figure}[h]
\begin{lstlisting}
(:action fill
        :parameters 
        (?r - robot ?c - cup 
         ?f - faucet ?l - location)
        :precondition 
        (and (is_grasped ?c) 
             (is_empty ?c) 
             (faucet_at ?f ?l) 
             (is_on ?f) 
             (robot_at ?r ?l))
        :effect 
        (and (is_filled ?c) 
             (not (is_on ?f)) 
             (is_off ?f))
)
\end{lstlisting}
\caption{Definition of action ``\emph{fill}'' in our PDDL-based task planner}\label{fig:fill1}
\end{figure}

The problem file, on the other hand, defines the task by specifying an initial state and a goal state, which in this case is ``Water is served to the user''. 
To generate a task plan, a solver, such as Fast Downward~\cite{helmert2006fast} can be used.
A feasible closed-world plan for this task is shown in Fig.~\ref{fig:plan1}.

\begin{figure}[h]
\begin{adjustwidth}{0.3cm}{0cm}
\begin{lstlisting}
S1: find robot cup kitchen
S2: find_faucet robot faucet kitchen
S3: turnon robot faucet kitchen
S4: grasp robot cup kitchen
S5: fill robot cup faucet kitchen
S6: move robot cup kitchen dining
S7: place robot cup table dining
\end{lstlisting}
\end{adjustwidth}
\caption{One closed-world plan for the task ``Serve water''}\label{fig:plan1}
\end{figure}

Plan Monitor is responsible for checking action feasibility under a given situation, such as ``Cup is dirty''. 
If an action, like \emph{fill}, is considered infeasible by an LLM, a constraint, such as \texttt{not (is\_dirty ?c)}, will be added to the action precondition. 
We implement a predicate generator that translates the natural language situation into a symbolic form, such as \texttt{is\_dirty ?cup}. 
This generator utilizes the few-shot learning capabilities of the LLM. 
Once the predicate is obtained, it is added to the PDDL code and highlighted with an underline, as shown in Fig.~\ref{fig:fill2}. 
Additionally, the initial state in the problem file is updated to reflect the new situation, \texttt{(is\_dirty cup)}.


\begin{figure}
\begin{lstlisting}
(:action fill
        :parameters 
        (?r - robot ?c - cup 
         ?f - faucet ?l - location)
        :precondition 
        (and (is_grasped ?c) 
             (is_empty ?c)
             |\underline{(not (is\_dirty $?$c))}|
             (faucet_at ?f ?l) 
             (is_on ?f) 
             (robot_at ?r ?l))
        :effect 
        (and (is_filled ?c) 
             (not (is_on ?f)) 
             (is_off ?f))
)
\end{lstlisting}
\caption{An action constraint, \texttt{not (is\_dirty ?c)}, is added into the action ``\emph{fill}''}\label{fig:fill2}
\end{figure}

After adding the constraint \texttt{not (is\_dirty ?c)}, the planning system might not generate a feasible plan, as the object \emph{cup} is found to be dirty and cannot be used. 
In such cases, the planning system needs to seek alternative solutions to accomplish the task by \emph{adding action effects} to the planner.
The commonsense knowledge that ``Bowl can be a container to fill water'' is obtained by Knowledge Acquirer.
This knowledge might have been overlooked by the planning system developer. 
COWP can add this new action effect into the planning system, as shown in Fig.~\ref{fig:fill3}. 
Now, the \emph{fill} action can be applied to both \emph{cup} and \emph{bowl}.
Additionally, related parts of the code are adjusted accordingly, and the initial state in the problem file is updated to include the bowl's location.

\begin{figure}
\begin{lstlisting}
(:action fill
        :parameters 
        (?r - robot ?b - bowl 
         ?f - faucet ?l - location)
        :precondition 
        (and (is_grasped ?b) 
             (is_empty ?b) 
             (faucet_at ?f ?l) 
             (is_on ?f) 
             (robot_at ?r ?l))
        :effect 
        (and (is_filled ?b) 
             (not (is_on ?f)) 
             (is_off ?f))
)
\end{lstlisting}
\caption{The PDDL-based task planner incorporates the commonsense knowledge that ``Bowl can be a container to fill water'' as an action effect.}\label{fig:fill3}
\end{figure}

An alternative plan that uses a bowl for serving water, shown in Fig.~\ref{fig:plan2}, can complete the service task ``Serve water'' and handle the situation ``Cup is dirty''.

\begin{figure}[h]
\begin{adjustwidth}{0.3cm}{0cm}
\begin{lstlisting}
S1: find robot bowl kitchen
S2: find_faucet robot faucet kitchen
S3: turnon robot faucet kitchen
S4: grasp robot bowl kitchen
S5: fill robot bowl faucet kitchen
S6: move robot bowl kitchen dining
S7: place robot bowl table dining
\end{lstlisting}
\end{adjustwidth}
\caption{A feasible plan for the task ``Serve water'', which can complete
the service task ``Serve water'' and handle the situation ``Cup is dirty.''}\label{fig:plan2}
\end{figure}

\section{Experiments}
In this section, we evaluate COWP's performance in planning and situation handling.

\subsection{Experimental Setup} 
Our experiments were performed in a \textit{dining domain}, where a service robot is tasked with fulfilling a user's service requests in a home setting. 
For simulating dining domains, we chose 12 everyday tasks from the \emph{ActivityPrograms} dataset~\cite{puig2018virtualhome}.
These tasks can be found in Table~\ref{tab:12tasks}.
For each task, we constructed PDDL-based planning systems to generate action sequences for their completion. 

To carry out our experiments, we used \mbox{OpenAI's} GPT-3 engines.
Please refer to Table~\ref{table:parameter} for the specific hyperparameters we adopted.
In simulation experiments, we assume that our robot is provided with a perception module for converting raw sensory data into logical facts (situations), such as objects and their properties, while the robot still needs to reason about the facts for planning and situation handling.
Our experiments were performed using simulated robot behaviors and situations collected from human participants. 


\begin{table}[t]
\centering
\caption{12 tasks extracted from the ActivityPrograms dataset~\cite{puig2018virtualhome} for evaluation purposes.}\label{tab:12tasks}
\begin{tabular}[t]{@{}l|l|l@{}}
\toprule
	 \textbf{Task Name} & \textbf{Task Name} & \textbf{Task Name}  \\ \midrule \midrule
	 Set table & Serve water & Serve coke \\ \midrule
      Wash plate & Heat burger & Make coffee \\ \midrule
	 Clean floor & Prepare burger & Store food \\ \midrule
      Wash cup & Wash sink & Wash glass\\
\bottomrule
\end{tabular}
\end{table}

\subsection{Simulation Platform}

\vspace{.5em}
\noindent
\textbf{Situation Dataset:}
To evaluate the performance of COWP in dealing with situations in a dining domain, 
we collected a dataset of execution-time situations using Amazon Mechanical Turk.
Each instance of the dataset corresponds to a situation that prevents a service robot from completing a task. 
We recruited MTurkers with a minimum HIT score of 70. 
Each MTurker was provided with a task description, including steps for completing the task. 
The MTurkers were asked to respond to a questionnaire by identifying one step in the provided plan and describing a situation that might occur in that step. 
For example, in the task ``Serve water'', we provided its plan consisting of the following steps: ``1) Walk to kitchen room, 2) Find glass, 3) Find sink, 4) Find faucet, 5) Turn on faucet, 6) Fill glass with water, 7) Move glass near table, 8) Place cup on table''.
Two common responses from the MTurkers were ``\emph{Glass is broken}'' for Step 2 and ``\emph{Faucet has no water}'' for Step 5.

\begin{table}[t]
\caption{Hypermeters of OpenAI's GPT-3 engines in Our Experiment}\label{table:parameter}
\begin{tabular}{l|c}
\toprule
\textbf{Parameter} & \textbf{Value} \\
\midrule
Model & text-davinci-003 \\
\midrule
Temperature & 0.0  \\
\midrule
Top p & 1.0 \\
\midrule
Maximum length & 32 \\
\midrule
Frequency penalty & 0.0 \\
\midrule
Presence penalty & 0.0 \\
\bottomrule
\end{tabular}
\end{table}

We received 1,224 responses from MTurkers, and \num~of them were valid, where the responses were evaluated through both a validation question and a manual filtering process.
We included a validation question that resembled other questions in the middle of the questionnaire, but instead of asking for a situation, it required the MTurker to input a provided ``secret code'', such as ``Successfully Verified!''.
Regarding the manual filtering process, we checked the responses for relevance to the task and logical coherence. 
For instance, some MTurkers copied and pasted text that was completely irrelevant, and those responses were removed manually. 
Additionally, while some responses like ``Glass's color is green'' were related to the task, such as ``Serve water,'' they were considered invalid because the robot can still complete the task regardless of the glass's color.
There are at least 88 situations collected for each of the 12 tasks. 
To facilitate the construction of the simulation platform, we further grouped the situations of significant similarities and generate a set of ``distinguishable'' situations.
For instance, two situations, ``Glass is broken'' and ``Glass is shattered'' are considered indistinguishable.
As a result, each task has 12 to 24 distinguishable situations, and the specific number of distinguishable situations for each task can be found in Table~\ref{tab:num_situations}.
Here, we show 15 distinguishable situations for the task ``Serve water'':

\begin{flushleft}
\item \hspace{0.5em} 1) \emph{Glass is broken.} 
\item \hspace{0.5em} 2) \emph{Faucet has no water.}
\item \hspace{0.5em} 3) \emph{Glass is dusty.} 
\item \hspace{0.5em} 4) \emph{Glass is missing.}
\item \hspace{0.5em} 5) \emph{Water is dirty.} 
\item \hspace{0.5em} 6) \emph{Faucet has sustained physical damage.}
\item \hspace{0.5em} 7) \emph{Faucet cannot be turned on.}
\item \hspace{0.5em} 8) \emph{Sink is not found.}
\item \hspace{0.5em} 9) \emph{Faucet is leaking.}
\item \hspace{0.5em} 10) \emph{Faucet is not found.}
\item \hspace{0.5em} 11) \emph{Water spills on floor.}
\item \hspace{0.5em} 12) \emph{Glass is not full of water.}
\item \hspace{0.5em} 13) \emph{Kitchen door is locked and cannot be opened.}
\item \hspace{0.5em} 14) \emph{Glass falls onto floor.}
\item \hspace{0.5em} 15) \emph{Glass is stuck and cannot be removed.}
\end{flushleft}
\vspace{0.5em}

On our project website (\url{https://cowplanning.github.io/}), users can download both the MTurk questionnaire and the situation dataset. 
The dataset is provided as a CSV file comprising 12 separate sheets, each representing the situations for a distinct task. 
The name of the task corresponds to the sheet name. 
Each sheet consists of five columns. 
The situations' descriptions provided by the MTurkers are in Column A. 
Column B details the corresponding steps where the described situation occurs. 
Column C is the index of distinguishable situations, while Column D provides descriptions of these situations. 
Finally, Column E indicates the number of distinguishable situations.

Note that \textbf{this dataset is intended solely for evaluation purposes}. The LLMs utilized in COWP act as zero-shot learners, and Plan Monitor can successfully perform its intended function without situation examples.

\begin{table}[t]
\centering
\caption{The number of distinguishable situations for each task.}\label{tab:num_situations}
\begin{tabular}[t]{@{}lc|lc@{}}
\toprule
	 \textbf{Task Name} & \textbf{Num} & \textbf{Task Name} & \textbf{Num} \\ \midrule \midrule
	 Make coffee  & 24 & Set table & 16 \\ \midrule
      prepare burger & 21 & Clean floor & 15 \\ \midrule
      Heat burger  & 19 & Serve water & 15\\ \midrule
	 Wash sink  & 18 & Wash cup  & 15 \\ \midrule
      Store food  & 17 & Serve coke  & 14 \\ \midrule
      Wash plate  & 16 & Wash glass  & 12\\
\bottomrule
\end{tabular}
\end{table}

\vspace{.5em}
\noindent
\textbf{Simulator:}
We have constructed a simulator capable of generating a high-level task name (e.g., ``Serve water'') and an unexpected situation (e.g., ``Glass is broken'') that occurs during one of the steps (e.g., ``Fill glass with water'') in each trial. 
Here, we assume that the probability of the robot encountering a situation while executing an action is denoted by $P$ (i.e., 0.1). 
This assumption implies that a longer task plan may be more likely to encounter such situations.
The simulator also contains 86 objects, including cups, burgers, forks, tables, and chairs. 
In each trial, it randomly selects and spawns half of the available objects for the robot to manipulate in order to resolve the situation.
This setting helps create a diverse environment for the robot.
These objects are also extracted from the ActivityPrograms dataset~\cite{puig2018virtualhome} and can be classified into five categories: kitchenware, appliance, furniture, food, and drink.
The ``kitchenware'' category comprises the highest number of objects (29), while the ``drink'' category contains the fewest (8).
For more details, please visit our website at \url{https://cowplanning.github.io/}.



\begin{figure*}[t]
\centering
\includegraphics[width=1.9\columnwidth]{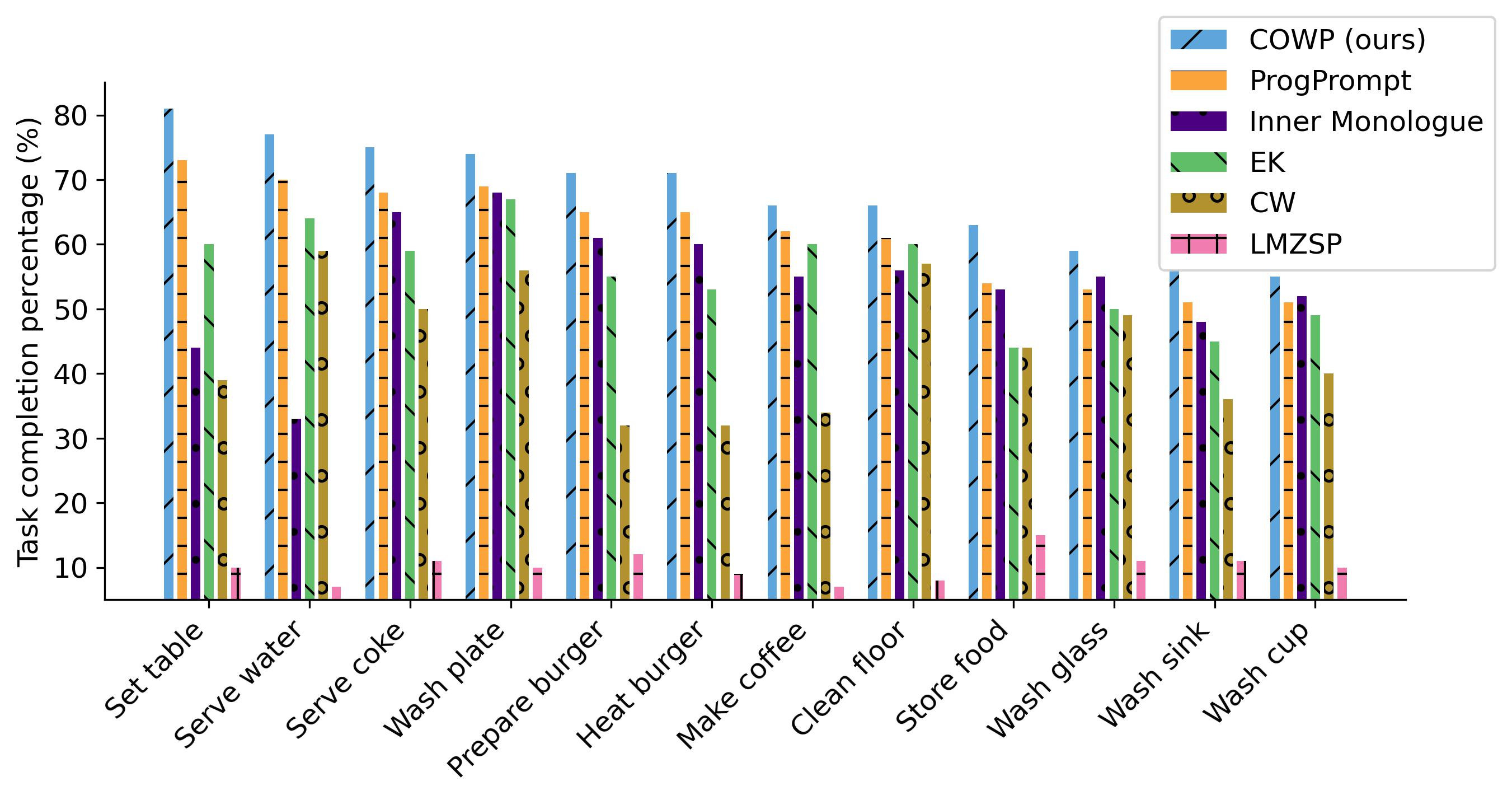}
\caption{The task completion percentage of COWP (ours) and five baseline methods under \textbf{12 different tasks}. The \textit{x-axis} represents the task name, and the \textit{y-axis} represents the task completion percentage.
The task completion percentage for each value is an average of 150 trials. 
The tasks are sorted based on the performance of COWP, where the very left corresponds to its best performance. 
}\label{fig:COWP_vs_baseline_task_completion}
\end{figure*}

\begin{table*}[htp]
\caption{The overall performances of COWP (ours) and five baseline methods are compared in terms of their task completion and situation handling percentages.
The task completion percentage is calculated by dividing the number of completed tasks by the total number of trials, which is 150.
The situation handling percentage is the ratio of the number of handled situations to the total number of situations.
}\label{table:compare}
\scriptsize
\centering
\begin{tabular}{l|c|c|c|c|c|c}
\toprule
\textbf{} & \textbf{COWP (ours)} & \textbf{\prog} & \textbf{\inner} & \textbf{EK} & \textbf{CW}  & \textbf{LMZSP} \\\midrule
\begin{tabular}[c]{@{}l@{}}\textbf{Task completion} \\ \textbf{percentage (\%)}\end{tabular} & \textbf{67.8} & 61.8 & 54.0 & 55.1  & 44.3 & 10.2\\\midrule
\begin{tabular}[c]{@{}l@{}}\textbf{Situation handling} \\ \textbf{percentage (\%)}\end{tabular} & \textbf{36.9} & 30.1 & 22.1 & 17.3 & 0.0 & 0.0\\
\bottomrule
\end{tabular}
\end{table*}

\begin{figure*}[t]
\centering
\includegraphics[width=1.7\columnwidth]{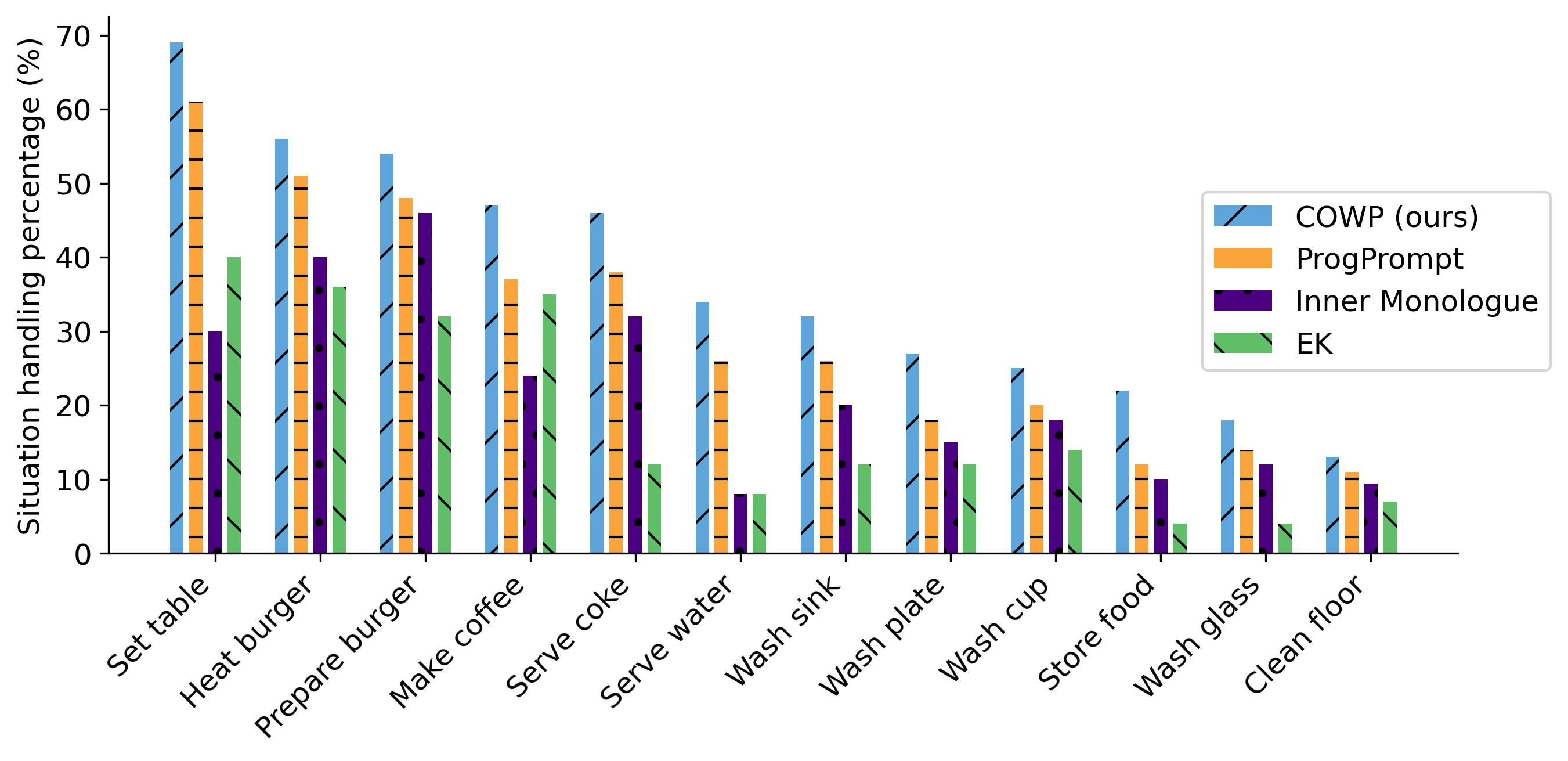}
\caption{The situation handling percentage of COWP (ours) and three baseline methods under \textbf{12 different tasks}, where the \textit{x-axis} represents the task name, and the \textit{y-axis} represents the situation handling percentage.
Each $y$ value represents a ratio of the number of handled situations to the total number of situations.
The tasks are ranked based on the performance of COWP, where the very left corresponds to its best performance. 
}\label{fig:COWP_vs_baseline_situation_handling}
\end{figure*}

\vspace{.5em}
\noindent
\textbf{Baselines and Evaluation Metrics}:
The evaluation of open-world planners is based on the respective \textit{success rates} of a robot completing service tasks and handling situations in a dining domain.
The following five baselines have been used in our experiments: 
\vspace{0.5em}
\begin{itemize}
    \item \inner~\cite{huang2022inner} leverages environmental feedback to generate task plans and handle situations. 
    In the original implementation, this approach could process three types of textual feedback: Passive Scene Description, Success Detection, and Active Scene Description. 
    Passive Scene Description involves obtaining feedback, such as details regarding available objects and situational context, from the environment. 
    One instance of this feedback could be ``cup,  the cup is broken, and drinking glass.'' 
    Success Detection checks if every action in the generated plan is successfully executed. 
    However, we have deactivated Active Scene Description because our system excludes humans from the loop.
    The specific prompts used in \inner~are available in the supplementary material on our project website.
    \vspace{.5em}
    \item Closed World (CW) corresponds to classical task planning developed for closed-world scenarios.
    In practice, CW was implemented by repeatedly activating the closed-world task planner and updating the current world state after executing each action. 
    CW does not have the capability to handle situations.
    For example, CW is unable to generate a plan that involves using a bowl as a container for serving water.
    \vspace{.5em}
    \item External Knowledge (EK)~\cite{jiang2019open} is a baseline approach that enables a closed-world task planner to acquire knowledge from an external source. 
    In our implementation, this external source provides information about a half of the domain objects.
    For instance, EK may generate a plan that involves using a bowl as a container for serving water, if its knowledge base contains ``Bowl can be used for serving water''.
    \vspace{.5em}
    \item Language Models as Zero-Shot Planners (\huang)~\cite{huang2022language} is a baseline that leverages LLM to compute task plans, where domain-specific action knowledge is not utilized. 
    The \huang~baseline~\cite{huang2022language} was not grounded. 
    As a result, their generated plans frequently involve objects unavailable or inapplicable in the current environment.
    Furthermore, \huang~cannot receive feedback from its environment, restricting its capability to handle situations.
    The hyperparameters used in LMZSP are available in the provided supplementary material on our project website.
    \vspace{.5em}
    \item \prog~\cite{singh2023progprompt} serves as the baseline method that utilizes a programmatic LLM prompt structure to facilitate open-world task planning. 
    The method consists of two types of prompts, namely PROMPT for Planning and PROMPT for State Feedback.
    \prog~is a few-shot learning method, unlike \huang.
    This is a \emph{competitive} baseline.
    More specifically, \prog~relies on the ``PROMPT for Planning'' to create a plan based on the task specification, including a contextual description of the situation. 
    The plan takes feedback from the environment into account by including preconditions for actions. 
    For instance, it might include situated state feedback like ``assert (`dirty' to `cup') else: find(`drinkingglass')''. 
    If the state is ``cup is dirty'', the ``find drinking glass'' action will be executed. 
    The `State Feedback Prompt' uses feedback from the environment to trigger preconditions in the generated plan. 
    In this example, a context description like ``cup is dirty'' would trigger the ``assert (`dirty' to `cup')'' precondition. 
    This is a competitive baseline. 
    \prog~has the same access to feedback inputs from the environment as our proposed approach, ensuring a fair comparison between the two methods. 
    For more detail on the exact prompts we used in our research with \prog, please refer to the provided supplementary materials on our project website.
\end{itemize}

Baselines \prog,~\inner~and \huang~have utilized the GPT-3 in their experimental implementations. 
It is crucial to note that GPT-3 offers multiple models, including ``\emph{text-davinci-002}'' or ``\emph{text-davinci-003}''. 
However, the specific model used by these studies was not mentioned. 
To ensure a fair comparison, we have chosen to use the ``\emph{text-davinci-003}'' model for all approaches.

\vspace{.5em}
\noindent
\textbf{Success Criteria:}
Unlike many classical planning systems~\cite{lo2020petlon,jiang2019open,garrett2021integrated,garrett2020pddlstream} that are guaranteed to provide sound solutions, LLM-based planning systems might generate invalid solutions without being aware of them.
For instance, GPT-3 might suggest that one can use a pan for drinking water, which is technically possible, but very uncommon in our everyday life. 
In this paper, we intend to consider those to be unsuccessful trials.
LLM-based task planners might wrongly believe that a good-quality plan is generated, while it is actually not the case. 
To compare with the ground truth, we recruited a second group of people, including six volunteers, to evaluate the performance of different open-world task planners. 
The volunteers included two females and four males aged 20-40, and all were graduate students in engineering fields. 
A successful task plan is defined as one that fulfills a user’s service request through actions that the robot is capable of executing and are acceptable to humans. 
In addition, common factors that could render a plan unsuccessful include, but are not limited to, the inclusion of non-existent objects in the environment, actions that the robot is unable to execute, or missing essential steps between actions. 
Detailed instructions provided for the volunteers prior to the evaluation are available in the provided supplementary materials on our project website.

\begin{figure*}[t]
\centering
\includegraphics[width=2.1\columnwidth]{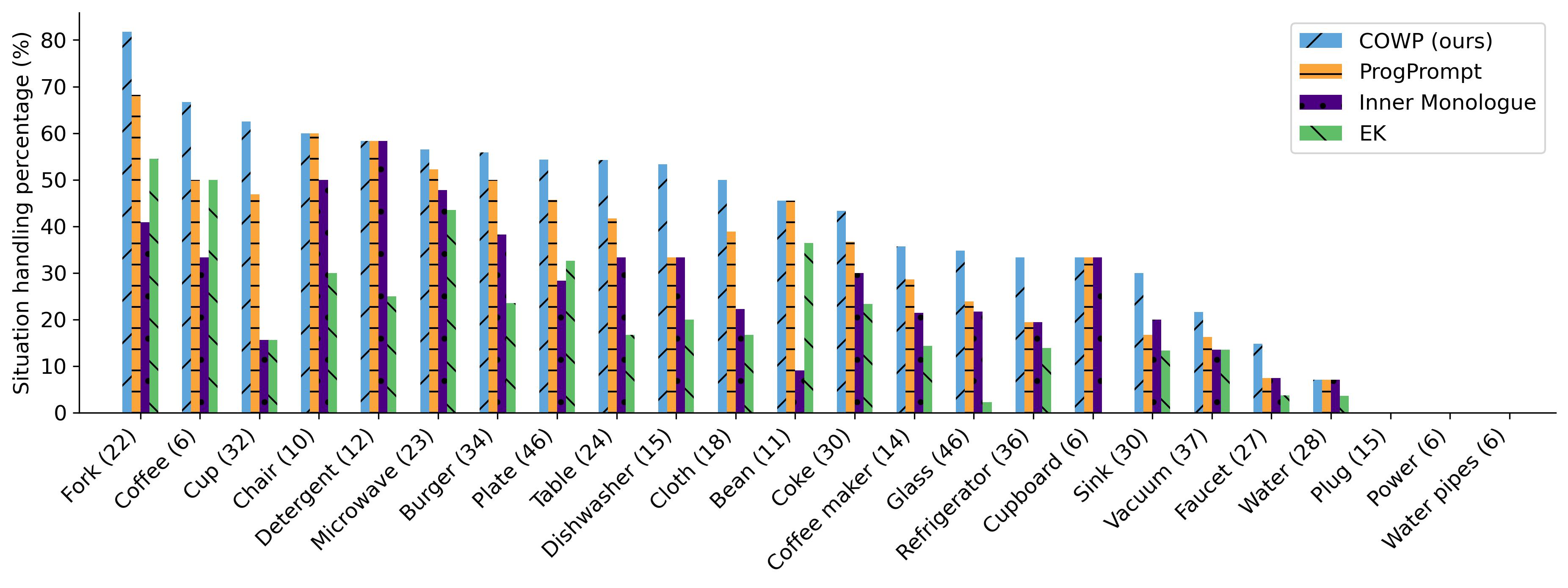}
\caption{The situation handling percentages of COWP (ours) and three baseline methods under \textbf{different objects}, where the \textit{x-axis} represents the object involved in the sampled situation, the number (X) beside each object is the occurrence of the object in situations, and the \textit{y-axis} represents the percentage of situation handling.
The objects are ranked based on the performance of COWP.
In this analysis, we only display objects with an occurrence of more than 5.
}\label{fig:objects}
\end{figure*}

\vspace{.5em}
\noindent
\textbf{COWP vs. Baselines (Overall):}
Table~\ref{table:compare} shows the overall performance of COWP in task planning and situation handling, as compared to five baselines, i.e., \prog, \inner, EK, \huang, and CW. 
The table demonstrates that COWP outperforms all five baselines in terms of task completion and situation handling percentages. 
However, \huang~and CW show poor performance. 
We believe the poor performance of CW is caused by its inability to leverage external information to handle unforeseen situations. 
For~\huang, its poor performance is caused by its weakness in grounding general commonsense knowledge in specific domains and the big noise in the generated plans.
As a result, many ``solutions'' generated by \huang~are not executable by the robot. 
For instance, in one plan, ``Find cup; Grab cup; Walk to sink; Run to cup; Walk to water; Find water; Grab water; Pour water into cup'' generated by \huang, some essential steps were omitted, such as ``Walk to kitchen'' and ``Turn on faucet''. 
Additionally, the robot was unable to execute the ``Grab water'' action.
This limitation of \huang~has also been observed and analyzed in the paper by Singh et al.~\cite{singh2023progprompt}.

The performance of \prog~is is not as good as that of COWP. 
This can be attributed to the fact that the task plans produced by \prog~might include noise, such as logical flaws, which reduces the chance of successfully completing the task. 
For example, one plan generated by \prog~includes the two consecutive actions of ``fill glass with water'' and ``turn on faucet'', which is invalid, because ``turn on faucet'' should be executed before ``fill glass with water.''
Additionally, we find the \prog's capability relies on the provided examples in the prompt, while COWP (ours) handles situations via zero-shot prompting of LLM. 

\inner's performance falls short when compared to COWP (ours) and ProgPrompt. 
Our observation indicates that \inner~tends to generate incomplete task plans and exhibit a bias toward ``Bounding Thinking,'' which consequently leads to a decrease in both task completion and situation handling percentages. 
Take, for instance, an \inner-produced task plan for serving water: ``Step 0: walk to kitchen; Step 1: find drinking glass; Step 2: fill drinking glass with water; Step 3: find kitchen table; Step 4: put drinking glass on kitchen table; Step 5: done''. 
This plan omits the crucial step of grasping the drinking glass first before filling it with water (Step 2), and overlooks the necessary action to access water, like ``Switch on Faucet'', given the environment does not have water readily available. 
This lack of completeness compromises the robot's ability to execute the task. 
Furthermore, \inner~has been shown to be lacking in situation handling. 
Despite being informed about situations through Passive Scene Description - a method of gathering environmental context, it often disregards this crucial contextual information. 
For instance, \inner~may recommend using a cup to serve water even if the ``cup is dirty.'' 
Moreover, it struggles to identify suitable solutions to address the situation. 
In the same scenario where a dirty cup is present, \inner~might suggest using a water boiler or a bucket to serve water to humans, even when more suitable options, such as a mug or a clean drinking glass, are available in the environment.

\vspace{.5em}
\noindent
\textbf{COWP vs. Baselines by Task:}
Fig.~\ref{fig:COWP_vs_baseline_task_completion} shows the results of comparing COWP and five baselines in the success rate of task completion under \emph{different tasks}.
The $x$-axis denotes the task name, and the $y$-axis indicates the percentage of task completion.
From the figure, we can see COWP outperforms the five baselines in all tasks.
It is quite interesting to see that open-world planners (including COWP) work better in some tasks than the others. 
For instance, in the ``Set table'' task, there are many ``missing object'' situations, and it happened that the robot could easily find alternative tableware objects to address those situations with the assistance of GPT-3. 
However, some tasks such as ``Wash glass'' are more difficult because many situations are beyond the robot's capabilities, e.g., ``There is a power outage'' and ``Faucet has no water''. 

Fig.~\ref{fig:COWP_vs_baseline_situation_handling} shows the results of comparing COWP and three baselines in the success rate of situation handling under \emph{different tasks}.
In this case, we do not include the baseline CW into the figure, because it’s inapplicable to the task of situation handling. 
The results indicate that COWP outperforms the baselines in all tasks.
Specifically, COWP was able to handle over 45\% of the situations in the first five tasks, and achieved the best performance of around 70\% in the ``Set table'' task. 
There are some tasks where situation handling is more difficult for COWP. 
For instance, there are situations, such as ``There is no water in the sink to wash the plate.'', and ``The robot cannot access the sink, as the door to the kitchen is locked.'' in task ``Wash plate'', where the robot could not do anything with its provided skills. 
By analyzing Fig.~\ref{fig:COWP_vs_baseline_task_completion} and Fig.~\ref{fig:COWP_vs_baseline_situation_handling}, we observe that the task ``Wash plate'' has a significantly higher task completion percentage than ``Wash glass'', despite having a similar situation handling percentage. 
This is due to ``Wash plate'' comprising only eight action sequences, compared to ``Wash glass'' with 12. 
Consequently, the robot is more prone to encountering situations in the ``Wash glass'' task, which hinders the robot’s task completion rate.

\begin{table*}[htp]
\caption{The performances of COWP (ours) and three baseline methods (\prog, \inner, and LMZSP) are compared in terms of their task completion and situation handling percentages. 
The comparisons are made using different LLM models (text-davinci-003, text-davinci-002, and text-ada-001) for two service tasks.}
\label{tab:llms}
\scriptsize
\centering
\begin{tabular}{l|c|c|c|c|c}
\toprule
\textbf{Task: Wash Glass} & \textbf{LLM Models} & \textbf{COWP (ours)} & \textbf{\prog} & \textbf{\inner} & \textbf{LMZSP} \\
\midrule
\multirow{3}{*}{\begin{tabular}[c]{@{}l@{}}Task completion \\ percentage (\%)\end{tabular}} & {text-davinci-003} & \textbf{59.3} & 53.3 & 54.7 & 10.7\\
& {text-davinci-002} & \textbf{53.7} & 36.0 & 27.3 & 5.3\\
& {text-ada-001} & \textbf{52.0} & 0.0 & 0.0 & 0.0\\
\midrule
\multirow{3}{*}{\begin{tabular}[c]{@{}l@{}}Situation handling \\ percentage (\%)\end{tabular}} & {text-davinci-003} & \textbf{18.0} & 14.0 & 12.0 & 0.0\\
& {text-davinci-002} & \textbf{14.0} & 6.0 & 4.0 & 0.0\\
& {text-ada-001} & \textbf{6.0} & 0.0 & 0.0 & 0.0\\
\midrule
\midrule
\textbf{Task: Serve Water} & \textbf{LLM Models} & \textbf{COWP (ours)} & \textbf{\prog} & \textbf{\inner} & \textbf{LMZSP} \\
\midrule
\multirow{3}{*}{\begin{tabular}[c]{@{}l@{}}Task completion \\ percentage (\%)\end{tabular}} & {text-davinci-003} & \textbf{76.6} & 69.3 & 32.7 & 6.7\\
& {text-davinci-002} & \textbf{73.3} & 15.4 & 13.3 & 2.7\\
& {text-ada-001} & \textbf{66.0} & 0.0 & 0.0 & 0.0\\
\midrule
\multirow{3}{*}{\begin{tabular}[c]{@{}l@{}}Situation handling \\ percentage (\%)\end{tabular}} & {text-davinci-003} & \textbf{34.7} & 26.5 & 8.2 & 0.0\\
& {text-davinci-002} & \textbf{30.6} & 8.2 & 6.1 & 0.0\\
& {text-ada-001} & \textbf{8.1} & 0.0 & 0.0 & 0.0\\
\bottomrule
\end{tabular}
\end{table*}

\begin{figure*}[t]
\centering
\includegraphics[width=2.0\columnwidth]{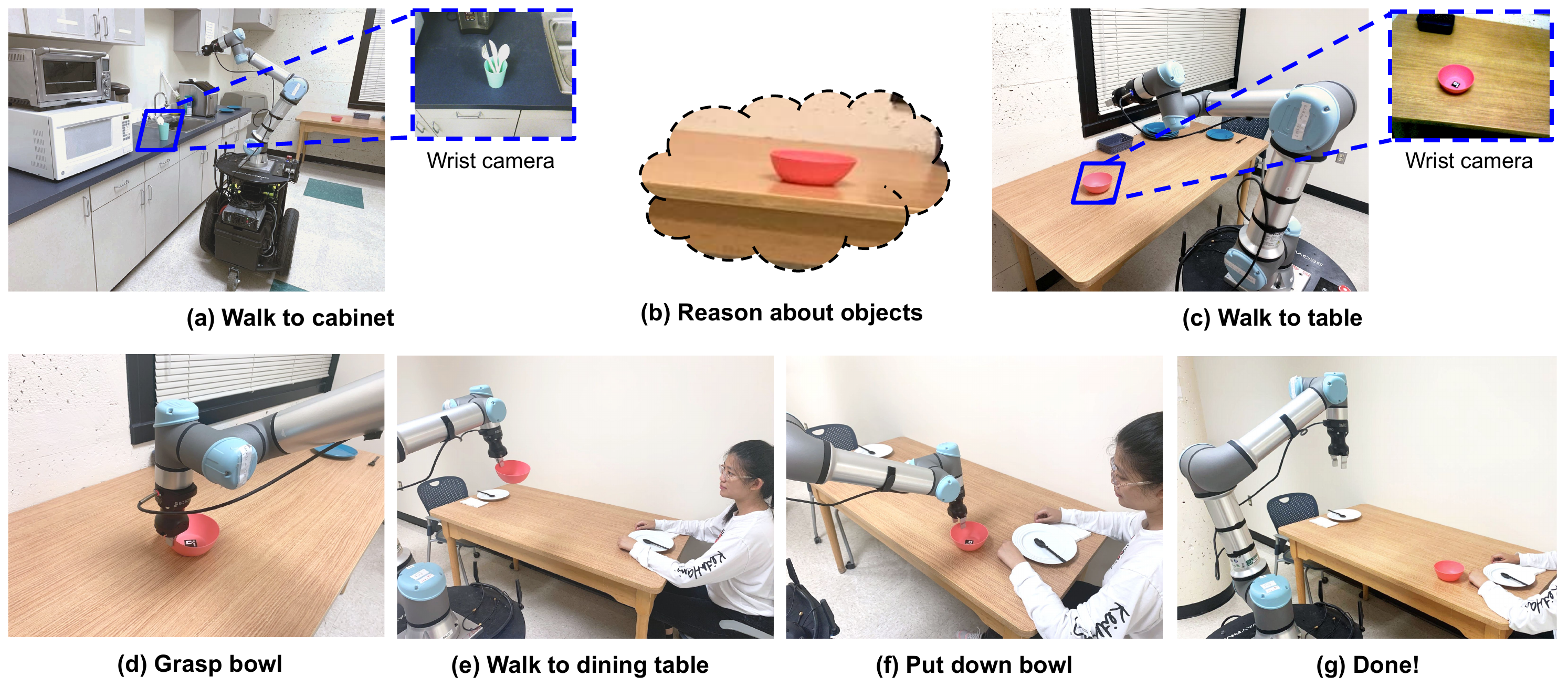}
\vspace{0.5em}
\caption{
An illustrative example of COWP for open-world planning, where the robot was tasked with ``delivering a cup for drinking water.'' 
\textbf{(a)} The robot walked to a cabinet, and located a cup on the cabinet. 
However, the robot found a situation that there were objects in the cup (a knife, a fork, and a spoon in this case). 
This observation was entered into the plan monitor, which queried GPT-3, and suggested that the planned action ``grasp'' was not applicable given the occupied cup. 
Accordingly, COWP updated its task planner by adding the new information that one cannot pour water into a non-empty cup. 
\textbf{(b)}~The robot reasoned about other objects that were available in the environment, and queried GPT-3 to update the task planner about whether those objects can be used for drinking water -- details in Section~\ref{sec:alg}. 
It happened that the robot learned a bowl could be used for drinking water. 
\textbf{(c)} A new plan of delivering a bowl to the human for drinking water was generated. 
Following the new plan, the robot walked to the table on which a bowl was located. 
\textbf{(d)} The robot grasped the bowl after observing it using vision. 
\textbf{(e)} The robot navigated to the dining table with the bowl. 
\textbf{(f)} The robot put down the bowl onto the dining table, and explained that a bowl was served due to the cup being occupied, which concluded the planning and execution processes. 
\textbf{(g)} The task is completed.
}\label{fig:real_demo}
\end{figure*}

\vspace{.5em}
\noindent
\textbf{COWP vs. Baselines by Object:}
Fig.~\ref{fig:objects} compares the performance of COWP and three baselines in situation handling percentage under \textit{different objects}.
The \textit{x-axis} represents the objects in our situation dataset and their occurrences, and the \textit{y-axis} represents the performance of planning methods in situation handling. 
For instance, common situations about cup include ``dirty cup,'' ``broken cup,'' and ``missing cup.'' 
COWP (ours) performed the best over all objects, while some baselines produced comparable performances over some objects. 
An important observation is that COWP produced higher success rates in situations involving ``fork'', ``coffee'', ``cup'', and ``chair''.
This is because many of those relevant situations are about missing objects, and the robot can easily find their alternatives in dining domains. 
By comparison, those situations involving ``power'', ``water pipes'', and ``plug'' are more difficult to the five methods (including COWP), because addressing those situations frequently requires skills beyond the robot's capabilities. 

\vspace{.5em}
\noindent
\textbf{COWP vs. Baselines under LLMs:}
Table~\ref{tab:llms} provides a comparison of COWP's performance with three baselines: \prog, \inner, and \huang. 
The evaluation specifically concentrates on the success rate of task completion and the ability to handle various situations under different LLMs. 
These baselines leverage LLMs for planning, while our method is classical planning augmented by LLMs. 
We selected three LLM models for this comparison: text-ada-001, text-davinci-002, and text-davinci-003. 
Text-ada–001 is considered to be the least effective, while text-davinci-003 is viewed as the most powerful~\cite{OpenAI2023}.
From the table, it is clear that the choice of LLM has a significant impact on the baseline models' performance, with all models performing at their best under text-davinci-003. 
The baselines face challenges in completing task planning when utilizing the text-ada-001 model. 
The choice of LLM also impacts the performance of COWP, particularly when using text-ada-001. 
We observe that the performance of COWP under text-ada-001 becomes poor. 
Even when there is a situation where a robot is unable to complete a task, COWP continues with its current actions. 
For example, COWP will still use a dirty cup to serve water to humans. 
Comparing our COWP with its baselines, it is clear that the choice of LLM has a more substantial effect on the baseline models, as they heavily rely on the selected LLMs in their planning phase.

\vspace{.5em}
\noindent
\textbf{Robot Demonstration: }
We demonstrated COWP using a mobile service robot that was tasked with delivering a cup for drinking water. 
The robot includes a UR5e arm, a Robotiq Hand-E gripper, and a Segway RMP-11 mobile base. 
The robot is capable of performing basic navigation and manipulation behaviors. 
Specifically, we employed GG-CNN~\cite{morrison2018closing} for object pick-and-place tasks. 
The navigation stack was built using the \textit{move\_base} package of the Robot Operating System (ROS)~\cite{quigley2009ros}. 
For visual scene analysis, the robot is equipped with a Robotiq Wrist Camera. 
With the assistance of Yolo-5~\cite{jocher2022ultralytics}, the robot recognizes objects in real time.
Upon receiving a top-down image, Yolo-5 generates the coordinates of the bounding boxes and identifies the objects within them, thus providing an understanding of the geometric relationships between objects. 
We employ a heuristics-based algorithm to determine whether the target object is occupied by another, such as when bounding boxes overlap, or if the target object is absent. 
While there exist more powerful tools, such as vision-language models, for visual scene analysis, computer vision is not our focus, and our goal here is to implement a basic perception component to close the perceive-reason-act loop.  

Fig.~\ref{fig:real_demo} shows a real-world demonstration where a robot used COWP for planning to complete the service task of ``Deliver a cup for drinking water.'' 
The initial plan included the following actions for the robot: 
\begin{enumerate}
    \item Walk to a cabinet on which a cup is located, 
    \item Grasp the cup after locating it, 
    \item Walk to dining table, and 
    \item Put down the cup to a table where the human is seated. 
\end{enumerate}

However, a situation was observed after executing Action~\#1, and the robot found \textit{the cup is occupied}. 
The robot initially did not know whether this affects its plan execution. 
After querying GPT-3, the robot learned that one cannot pour water into an occupied cup, which renders its current plan infeasible. 
COWP enabled the robot to reason about other objects of the environment. 
The robot iteratively queried GPT-3 about whether $X$ can be used for drinking water (using Prompt Template 2 in Section~\ref{sec:alg}), where $X$ is one of the objects from the environment. 
It happened that the robot learned ``bowl'' can be used for drinking water. 
With this newly learned, task-oriented commonsense knowledge, COWP successfully helped the robot generate a new plan, which used the bowl instead, to fulfill the service request. 

We have generated a demo video that has been uploaded as part of the supplementary materials. 

\section{Conclusion and Future Work}\label{sec:conclusion}

In this paper, we develop a Large Language Model-based open-world task planning system for robots, called COWP, towards robust task planning and situation handling in open worlds. 
The novelty of COWP points to the integration of a classical, knowledge-based task planning system, and a pretrained language model for commonsense knowledge acquisition. 
The marriage of the two enables COWP to ground domain-independent commonsense knowledge to specific task planning problems. 
To evaluate COWP systematically, we collected a situation dataset that includes \num~execution-time situations in a dining domain. 
Experimental results suggest that COWP performed better than existing task planners developed for closed-world and open-world scenarios. 
We also provided a demonstration of COWP using a mobile manipulator working on delivery tasks, which provides a reference to COWP practitioners for real-world applications. 

We recognize the merits and limitations of ``pure'' LLM for planning (e.g., \prog) and classical planning augmented with LLMs, particularly in terms of logical consistency and flexibility of representation. 
Pure LLM planning has the advantage of greater flexibility, allowing it to handle a wider range of unforeseen situations, but it may be more prone to logical errors. 
In contrast, LLM-augmented classical planning, by parsing the LLM outputs into structured formats using hand-crafted rules, can reduce the occurrence of logical errors but may be less flexible when dealing with unforeseen situations. 
For instance, there might be situations where our hand-crafted rules fail to parse the output from LLMs. 

It is evident that prompt design significantly affects the performance of Large Language Models (LLMs)~\cite{zhang2021differentiable}, which has motivated the recent research on prompt engineering. 
For instance, we tried replacing ``suitable'' with ``possible'' and ``recommended,'' in the prompt templates and observed a decline in the system performance. 
Researchers can improve the prompt design of COWP in future work. 
There is the potential that other LLMs, such as ChatGPT~\cite{openai} and Bard~\cite{google-bard-faq}, can produce better performance in open-world planning, and their performances might be domain-dependent, which can lead to very interesting future research. 
We present a complete implementation of COWP on a real robot, while acknowledging that there are many ways to improve the implementations. 
For instance, one can use a more advanced visual scene analysis tool to generate more informative observations for situation detection, or equip the robot with more skills (such as wiping a table,  moving a chair, and opening a door) to deal with situations that cannot be handled now. 

We acknowledge the limitations of our method in comprehensively understanding the world state in real-world systems. 
To effectively monitor plans, an autonomous system requires a general-purpose perception system capable of recognizing various situations, such as object states (dirty, broken, spilled, etc.) and event occurrences. 
While this task is challenging, recent developments in perception, such as vision-language models, hold promise in combining visual perception with language understanding to enhance autonomous systems' capability in detecting and interpreting situations~\cite{zhang2023grounding, zhu2023minigpt, gao2023llama}. 
We are optimistic about overcoming these challenges in our future work.

\bmhead{Acknowledgments}
A portion of this work has taken place at the Autonomous Intelligent Robotics (AIR) Group, SUNY Binghamton. AIR research is supported in part by grants from the National Science Foundation (NRI-1925044), Ford Motor Company, OPPO, and SUNY Research Foundation.


\section*{Declarations}

\begin{itemize}
\item Funding: A portion of this work has taken place at the Autonomous Intelligent Robotics (AIR) Group, SUNY Binghamton. AIR research is supported in part by grants from the National Science Foundation (NRI-1925044), Ford Motor Company, OPPO, and SUNY Research Foundation.
\item Conflict of interest: The authors declare that they have no conflict of interest.
\item Authors' contributions: Yan Ding, Xiaohan Zhang, Saeid Amiri, Hao Yang, Andy Kaminski, Chad Esselink, and Shiqi Zhang contributed to the development of the initial ideas and methodology.
Yan Ding, Xiaohan Zhang, and Saeid Amiri contributed to implementing the methodology.
Yan Ding, Xiaohan Zhang, Saeid Amiri, and Nieqing Cao contributed to the experiments.
Yan Ding, Xiaohan Zhang, Saeid Amiri, Hao Yang, and Shiqi Zhang contributed to the analysis of the results.
Yan Ding, Xiaohan Zhang, Saeid Amiri, and Shiqi Zhang contributed to the manuscript writing.
All authors reviewed and provided feedback on the manuscript.

\end{itemize}

\bibliography{sn-bibliography}

\begin{thebibliography}{10}
\providecommand{\doi}[1]{\url{https://doi.org/#1}}
\bibcommenthead

\bibitem[\protect\citeauthoryear{Ghallab et~al.}{2016}]{ghallab2016automated}
Ghallab M, Nau D, Traverso P.
\newblock Automated planning and acting.
\newblock Cambridge University Press; 2016.

\bibitem[\protect\citeauthoryear{Reiter}{1981}]{reiter1981closed}
Reiter R.
\newblock On closed world data bases.
\newblock In: Readings in artificial intelligence. Elsevier; 1981. p. 119--140.

\bibitem[\protect\citeauthoryear{Knoblock
  et~al.}{1991}]{knoblock1991characterizing}
Knoblock CA, Tenenberg JD, Yang Q.
\newblock Characterizing abstraction hierarchies for planning.
\newblock In: Proceedings of the ninth National conference on Artificial
  intelligence-Volume 2; 1991. p. 692--697.

\bibitem[\protect\citeauthoryear{Hoffmann}{2001}]{hoffmann2001ff}
Hoffmann J.
\newblock FF: The fast-forward planning system.
\newblock AI magazine. 2001;22(3):57--57.

\bibitem[\protect\citeauthoryear{Nau et~al.}{2003}]{nau2003shop2}
Nau DS, Au TC, Ilghami O, Kuter U, Murdock JW, Wu D, et~al.
\newblock SHOP2: An HTN planning system.
\newblock Journal of artificial intelligence research. 2003;20:379--404.

\bibitem[\protect\citeauthoryear{Helmert}{2006}]{helmert2006fast}
Helmert M.
\newblock The fast downward planning system.
\newblock Journal of Artificial Intelligence Research. 2006;26:191--246.

\bibitem[\protect\citeauthoryear{Hanheide et~al.}{2017}]{hanheide2017robot}
Hanheide M, G{\"o}belbecker M, Horn GS, Pronobis A, Sj{\"o}{\"o} K, Aydemir A,
  et~al.
\newblock Robot task planning and explanation in open and uncertain worlds.
\newblock Artificial Intelligence. 2017;247:119--150.

\bibitem[\protect\citeauthoryear{Jiang et~al.}{2019}]{jiang2019open}
Jiang Y, Walker N, Hart J, Stone P.
\newblock Open-world reasoning for service robots.
\newblock In: Proceedings of the International Conference on Automated Planning
  and Scheduling. vol.~29; 2019. p. 725--733.

\bibitem[\protect\citeauthoryear{Chernova et~al.}{2020}]{chernova2020situated}
Chernova S, Chu V, Daruna A, Garrison H, Hahn M, Khante P, et~al.
\newblock Situated bayesian reasoning framework for robots operating in diverse
  everyday environments.
\newblock In: Robotics Research. Springer; 2020. p. 353--369.

\bibitem[\protect\citeauthoryear{Kant et~al.}{2022}]{kant2022housekeep}
Kant Y, Ramachandran A, Yenamandra S, Gilitschenski I, Batra D, Szot A, et~al.
\newblock Housekeep: Tidying virtual households using commonsense reasoning.
\newblock In: Computer Vision--ECCV 2022. Springer; 2022. p. 355--373.

\bibitem[\protect\citeauthoryear{Huang et~al.}{2022}]{huang2022language}
Huang W, Abbeel P, Pathak D, Mordatch I.
\newblock Language Models as Zero-Shot Planners: Extracting Actionable
  Knowledge for Embodied Agents.
\newblock Thirty-ninth International Conference on Machine Learning. 2022;.

\bibitem[\protect\citeauthoryear{Brohan et~al.}{2023}]{brohan2023can}
Brohan A, Chebotar Y, Finn C, Hausman K, Herzog A, Ho D, et~al.
\newblock Do as i can, not as i say: Grounding language in robotic affordances.
\newblock In: Conference on Robot Learning; 2023. p. 287--318.

\bibitem[\protect\citeauthoryear{Perera et~al.}{2015}]{perera2015learning}
Perera V, Soetens R, Kollar T, Samadi M, Sun Y, Nardi D, et~al.
\newblock Learning task knowledge from dialog and web access.
\newblock Robotics. 2015;4(2):223--252.

\bibitem[\protect\citeauthoryear{Amiri et~al.}{2019}]{amiri2019augmenting}
Amiri S, Bajracharya S, Goktolgal C, Thomason J, Zhang S.
\newblock Augmenting knowledge through statistical, goal-oriented human-robot
  dialog.
\newblock In: 2019 IEEE/RSJ International Conference on Intelligent Robots and
  Systems (IROS). IEEE; 2019. p. 744--750.

\bibitem[\protect\citeauthoryear{Tucker et~al.}{2020}]{tucker2020learning}
Tucker M, Aksaray D, Paul R, Stein GJ, Roy N.
\newblock Learning unknown groundings for natural language interaction with
  mobile robots.
\newblock In: Robotics Research. Springer; 2020. p. 317--333.

\bibitem[\protect\citeauthoryear{Brown et~al.}{2020}]{brown2020language}
Brown T, Mann B, Ryder N, Subbiah M, Kaplan JD, Dhariwal P, et~al.
\newblock Language models are few-shot learners.
\newblock Advances in neural information processing systems.
  2020;33:1877--1901.

\bibitem[\protect\citeauthoryear{Zhang et~al.}{2022}]{zhang2022opt}
Zhang S, Roller S, Goyal N, Artetxe M, Chen M, Chen S, et~al.
\newblock OPT: Open Pre-trained Transformer Language Models.
\newblock arXiv preprint arXiv:220501068. 2022;.

\bibitem[\protect\citeauthoryear{OpenAI}{2023}]{openai}
OpenAI.: ChatGPT.
\newblock Cit. on pp. 1, 16.
\newblock Accessed: 2023-02-08.
\newblock Available from: \url{https://openai.com/blog/chatgpt/}.

\bibitem[\protect\citeauthoryear{{Google}}{2023}]{google-bard-faq}
{Google}.: Bard FAQ.
\newblock Accessed on April 7, 2023.
\newblock \url{https://bard.google.com/faq}.

\bibitem[\protect\citeauthoryear{Elsweiler et~al.}{2022}]{elsweiler2022food}
Elsweiler D, Hauptmann H, Trattner C.
\newblock Food recommender systems.
\newblock In: Recommender Systems Handbook. Springer; 2022. p. 871--925.

\bibitem[\protect\citeauthoryear{Davis and Marcus}{2015}]{davis2015commonsense}
Davis E, Marcus G.
\newblock Commonsense reasoning and commonsense knowledge in artificial
  intelligence.
\newblock Communications of the ACM. 2015;58(9):92--103.

\bibitem[\protect\citeauthoryear{Huang et~al.}{2023}]{huang2023grounded}
Huang W, Xia F, Shah D, Driess D, Zeng A, Lu Y, et~al.
\newblock Grounded Decoding: Guiding Text Generation with Grounded Models for
  Robot Control.
\newblock arXiv preprint arXiv:230300855. 2023;.

\bibitem[\protect\citeauthoryear{Puig et~al.}{2018}]{puig2018virtualhome}
Puig X, Ra K, Boben M, Li J, Wang T, Fidler S, et~al.
\newblock Virtualhome: Simulating household activities via programs.
\newblock In: Proceedings of the IEEE Conference on Computer Vision and Pattern
  Recognition; 2018. p. 8494--8502.

\bibitem[\protect\citeauthoryear{Singh et~al.}{2023}]{singh2023progprompt}
Singh I, Blukis V, Mousavian A, Goyal A, Xu D, Tremblay J, et~al.
\newblock Progprompt: Generating situated robot task plans using large language
  models.
\newblock International Conference on Robotics and Automation (ICRA). 2023;.

\bibitem[\protect\citeauthoryear{Haslum et~al.}{2019}]{haslum2019introduction}
Haslum P, Lipovetzky N, Magazzeni D, Muise C.
\newblock An introduction to the planning domain definition language.
\newblock Synthesis Lectures on Artificial Intelligence and Machine Learning.
  2019;13(2):1--187.

\bibitem[\protect\citeauthoryear{Jiang et~al.}{2019}]{jiang2019task}
Jiang Yq, Zhang Sq, Khandelwal P, Stone P.
\newblock Task planning in robotics: an empirical comparison of PDDL-and
  ASP-based systems.
\newblock Frontiers of Information Technology \& Electronic Engineering.
  2019;20(3):363--373.

\bibitem[\protect\citeauthoryear{Galindo et~al.}{2008}]{galindo2008robot}
Galindo C, Fern{\'a}ndez-Madrigal JA, Gonz{\'a}lez J, Saffiotti A.
\newblock Robot task planning using semantic maps.
\newblock Robotics and autonomous systems. 2008;56(11):955--966.

\bibitem[\protect\citeauthoryear{Valmeekam et~al.}{2022}]{valmeekam2022large}
Valmeekam K, Olmo A, Sreedharan S, Kambhampati S.
\newblock Large Language Models Still Can't Plan (A Benchmark for LLMs on
  Planning and Reasoning about Change).
\newblock arXiv preprint arXiv:220610498. 2022;.

\bibitem[\protect\citeauthoryear{Valmeekam
  et~al.}{2023}]{valmeekam2023planning}
Valmeekam K, Sreedharan S, Marquez M, Olmo A, Kambhampati S.
\newblock On the Planning Abilities of Large Language Models (A Critical
  Investigation with a Proposed Benchmark).
\newblock arXiv preprint arXiv:230206706. 2023;.

\bibitem[\protect\citeauthoryear{OpenAI}{2023}]{openai2023gpt4}
OpenAI.: GPT-4 Technical Report.

\bibitem[\protect\citeauthoryear{Liu et~al.}{2023}]{liu2023llm}
Liu B, Jiang Y, Zhang X, Liu Q, Zhang S, Biswas J, et~al.
\newblock LLM+P: Empowering Large Language Models with Optimal Planning
  Proficiency.
\newblock arXiv preprint arXiv:230411477. 2023;.

\bibitem[\protect\citeauthoryear{Devlin et~al.}{2018}]{devlin2018bert}
Devlin J, Chang MW, Lee K, Toutanova K.
\newblock Bert: Pre-training of deep bidirectional transformers for language
  understanding.
\newblock arXiv preprint arXiv:181004805. 2018;.

\bibitem[\protect\citeauthoryear{Chen et~al.}{2021}]{chen2021evaluating}
Chen M, Tworek J, Jun H, Yuan Q, Pinto HPdO, Kaplan J, et~al.
\newblock Evaluating large language models trained on code.
\newblock arXiv preprint arXiv:210703374. 2021;.

\bibitem[\protect\citeauthoryear{Liu et~al.}{2023}]{liu2023pre}
Liu P, Yuan W, Fu J, Jiang Z, Hayashi H, Neubig G.
\newblock Pre-train, prompt, and predict: A systematic survey of prompting
  methods in natural language processing.
\newblock ACM Computing Surveys. 2023;55(9):1--35.

\bibitem[\protect\citeauthoryear{Wang et~al.}{2021}]{wang2021can}
Wang C, Liu P, Zhang Y.
\newblock Can Generative Pre-trained Language Models Serve As Knowledge Bases
  for Closed-book QA?
\newblock In: Proceedings of the 59th Annual Meeting of the Association for
  Computational Linguistics and the 11th International Joint Conference on
  Natural Language Processing; 2021. p. 3241--3251.

\bibitem[\protect\citeauthoryear{Li et~al.}{2003}]{li2022pre}
Li S, Puig X, Paxton C, Du Y, Wang C, Fan L, et~al.
\newblock Pre-trained language models for interactive decision-making.
\newblock Advances in Neural Information Processing Systems. 2003;.

\bibitem[\protect\citeauthoryear{West et~al.}{2022}]{west2021symbolic}
West P, Bhagavatula C, Hessel J, Hwang JD, Jiang L, Bras RL, et~al.
\newblock Symbolic knowledge distillation: from general language models to
  commonsense models.
\newblock Proceedings of the 2022 Conference of the North American Chapter of
  the Association for Computational Linguistics: Human Language Technologies.
  2022;.

\bibitem[\protect\citeauthoryear{Huang et~al.}{2022}]{huang2022inner}
Huang W, Xia F, Xiao T, Chan H, Liang J, Florence P, et~al.
\newblock Inner Monologue: Embodied Reasoning through Planning with Language
  Models.
\newblock In: 6th Annual Conference on Robot Learning; 2022. .

\bibitem[\protect\citeauthoryear{Ding et~al.}{2023}]{ding2023task}
Ding Y, Zhang X, Paxton C, Zhang S.
\newblock Task and Motion Planning with Large Language Models for Object
  Rearrangement.
\newblock arXiv preprint arXiv:230306247. 2023;.

\bibitem[\protect\citeauthoryear{Xie et~al.}{2023}]{xie2023translating}
Xie Y, Yu C, Zhu T, Bai J, Gong Z, Soh H.
\newblock Translating natural language to planning goals with large-language
  models.
\newblock arXiv preprint arXiv:230205128. 2023;.

\bibitem[\protect\citeauthoryear{Song et~al.}{2022}]{song2022llm}
Song CH, Wu J, Washington C, Sadler BM, Chao WL, Su Y.
\newblock Llm-planner: Few-shot grounded planning for embodied agents with
  large language models.
\newblock arXiv preprint arXiv:221204088. 2022;.

\bibitem[\protect\citeauthoryear{Lin et~al.}{2023}]{lin2023text2motion}
Lin K, Agia C, Migimatsu T, Pavone M, Bohg J.
\newblock Text2Motion: From Natural Language Instructions to Feasible Plans.
\newblock arXiv preprint arXiv:230312153. 2023;.

\bibitem[\protect\citeauthoryear{Aeronautiques
  et~al.}{1998}]{aeronautiques1998pddl}
Aeronautiques C, Howe A, Knoblock C, McDermott ID, Ram A, Veloso M, et~al.
\newblock PDDL| The Planning Domain Definition Language.
\newblock Technical Report, Tech Rep. 1998;.

\bibitem[\protect\citeauthoryear{Lo et~al.}{2020}]{lo2020petlon}
Lo SY, Zhang S, Stone P.
\newblock The petlon algorithm to plan efficiently for task-level-optimal
  navigation.
\newblock Journal of Artificial Intelligence Research. 2020;69:471--500.

\bibitem[\protect\citeauthoryear{Garrett et~al.}{2021}]{garrett2021integrated}
Garrett CR, Chitnis R, Holladay R, Kim B, Silver T, Kaelbling LP, et~al.
\newblock Integrated task and motion planning.
\newblock Annual review of control, robotics, and autonomous systems.
  2021;4:265--293.

\bibitem[\protect\citeauthoryear{Garrett et~al.}{2020}]{garrett2020pddlstream}
Garrett CR, Lozano-P{\'e}rez T, Kaelbling LP.
\newblock Pddlstream: Integrating symbolic planners and blackbox samplers via
  optimistic adaptive planning.
\newblock In: Proceedings of the International Conference on Automated Planning
  and Scheduling. vol.~30; 2020. p. 440--448.

\bibitem[\protect\citeauthoryear{OpenAI}{}]{OpenAI2023}
OpenAI.: Models - OpenAI API.
\newblock Accessed: 2023-07-10.
\newblock \url{https://platform.openai.com/docs/models/overview}.

\bibitem[\protect\citeauthoryear{Morrison et~al.}{2018}]{morrison2018closing}
Morrison D, Corke P, Leitner J.
\newblock Closing the loop for robotic grasping: A real-time, generative grasp
  synthesis approach.
\newblock arXiv preprint arXiv:180405172. 2018;.

\bibitem[\protect\citeauthoryear{Quigley et~al.}{2009}]{quigley2009ros}
Quigley M, Conley K, Gerkey B, Faust J, Foote T, Leibs J, et~al.
\newblock ROS: an open-source Robot Operating System.
\newblock In: ICRA workshop on open source software. vol.~3. Kobe, Japan; 2009.
  p.~5.

\bibitem[\protect\citeauthoryear{Jocher et~al.}{2022}]{jocher2022ultralytics}
Jocher G, Chaurasia A, Stoken A, Borovec J, Kwon Y, Michael K, et~al.
\newblock ultralytics/yolov5: v7. 0-yolov5 sota realtime instance segmentation.
\newblock Zenodo. 2022;.

\bibitem[\protect\citeauthoryear{Zhang et~al.}{2021}]{zhang2021differentiable}
Zhang N, Li L, Chen X, Deng S, Bi Z, Tan C, et~al.
\newblock Differentiable Prompt Makes Pre-trained Language Models Better
  Few-shot Learners.
\newblock In: International Conference on Learning Representations; 2021. .

\bibitem[\protect\citeauthoryear{Zhang et~al.}{2023}]{zhang2023grounding}
Zhang X, Ding Y, Amiri S, Yang H, Kaminski A, Esselink C, et~al.
\newblock Grounding Classical Task Planners via Vision-Language Models.
\newblock arXiv preprint arXiv:230408587. 2023;.

\bibitem[\protect\citeauthoryear{Zhu et~al.}{2023}]{zhu2023minigpt}
Zhu D, Chen J, Shen X, Li X, Elhoseiny M.
\newblock Minigpt-4: Enhancing vision-language understanding with advanced
  large language models.
\newblock arXiv preprint arXiv:230410592. 2023;.

\bibitem[\protect\citeauthoryear{Gao et~al.}{2023}]{gao2023llama}
Gao P, Han J, Zhang R, Lin Z, Geng S, Zhou A, et~al.
\newblock Llama-adapter v2: Parameter-efficient visual instruction model.
\newblock arXiv preprint arXiv:230415010. 2023;.

\end{thebibliography}

\end{document}